%% file: main.tex
\pdfoutput=1

\documentclass[11pt]{article}

\usepackage[dvipsnames]{xcolor}

\usepackage[]{EMNLP2023}

\usepackage{times}
\usepackage{latexsym}

\usepackage[T1]{fontenc}

\usepackage[utf8]{inputenc}

\usepackage{microtype}

\usepackage{inconsolata}

\usepackage{booktabs}
\usepackage{multirow}
\usepackage{adjustbox}
\usepackage{amssymb}
\usepackage{pifont}
\usepackage[linewidth=0.5pt]{mdframed}

\newcommand{\cmark}{\ding{51}}
\newcommand{\xmark}{\ding{55}}

\title{Training and Meta-Evaluating Machine Translation Evaluation Metrics at the Paragraph Level}

\author{Daniel Deutsch, Juraj Juraska, Mara Finkelstein, and Markus Freitag \\
  Google \\
  \texttt{\{dandeutsch,jjuraska,marafin,freitag\}@google.com}
}

\begin{document}

\maketitle

\input{00_abstract}
\input{01_introduction}
\input{02_terminology}
\input{03_datasets}

\input{04_metrics}
\input{05_setup}
\input{06_results}
\input{07_discussion}

\input{08_related_work}
\input{09_conclusion}
\input{10_limitations}

\bibliography{custom}
\bibliographystyle{acl_natbib}

\appendix

\input{appendix/dataset_statistics}
\input{appendix/additional_results}

\end{document}

%% file: 00_abstract.tex
\begin{abstract}
As research on machine translation moves to translating text beyond the sentence level, it remains unclear how effective automatic evaluation metrics are at scoring longer translations.
In this work, we first propose a method for creating paragraph-level data for training and meta-evaluating metrics from existing sentence-level data.
Then, we use these new datasets to benchmark existing sentence-level metrics as well as train learned metrics at the paragraph level.
Interestingly, our experimental results demonstrate that using sentence-level metrics to score entire paragraphs is equally as effective as using a metric designed to work at the paragraph level.
We speculate this result can be attributed to properties of the task of reference-based evaluation as well as limitations of our datasets with respect to capturing all types of phenomena that occur in paragraph-level translations.
\end{abstract}

%% file: 01_introduction.tex
\section{Introduction}
\label{sec:intro}

Automatic evaluation metrics have always been a critical component to the progress of research on machine translation (MT).
As the field of MT moves beyond translating individual sentences to translating full paragraphs, book chapters, or documents \citep{10.1162/tacl_a_00029,sun-etal-2022-rethinking,thai-etal-2022-exploring,jiang-etal-2023-discourse,post2023escaping}, automatic metrics need to be designed to work on these longer texts.

Currently, how well automatic metrics agree with human judgments of paragraph translation quality is an open question.\footnote{
    Translation beyond the sentence level is often referred to as document-level MT.
    However, there is no clear definition for the term ``document.''
    We use ``paragraph'' in this work because we feel it most accurately describes the length of text in our datasets.
    See \S\ref{sec:terminology} for more details on this.
}
Few studies have meta-evaluated metrics on longer texts, and those that have are focused on the literary domain and are limited in the size of the evaluation dataset \citep{jiang-etal-2022-blonde,thai-etal-2022-exploring,karpinska2023large}.
In this work, we investigate training and meta-evaluating metrics for scoring paragraph translations using the benchmark Workshop on Machine Translation (WMT) datasets that are widely used for metric development \citep{WMT22}.

Due to the scarcity of human ratings of paragraph translations, we propose a method to create paragraph-level training and meta-evaluation datasets from the existing WMT sentence-level datasets (\S\ref{sec:datasets}).
Although these ratings are typically only used at the sentence level, they were collected on contiguous paragraphs and performed with document context, so they can be used as paragraph-level datasets.
We repurpose these datasets to benchmark existing sentence-level metrics as well as train new paragraph-level metrics for scoring paragraph translations (\S\ref{sec:para_level_metrics}).

Our experimental results are somewhat surprising.
We find that there appears to be little evidence that training on paragraph-level data is beneficial---at least given the limitations of our experimental setup.
Using metrics trained on sentence-level data only to directly score full paragraphs achieves comparable agreement to human ratings as metrics trained on paragraph-level data (\S\ref{sec:para_eval}).
Sentence-level metrics appear to generalize well to inputs much longer than they were trained on (\S\ref{sec:seg_comparison}).

We hypothesize these observations can be explained by the nature of evaluating translations and characteristics of our paragraph-level dataset (\S\ref{sec:discussion}).
We speculate that long range dependencies---which paragraph-level metrics can model but sentence-level likely do not---may not be too important for achieving high agreement with human ratings.
Further, due to the fact that our training and evaluation datasets assume a sentence alignment between the reference and hypothesis paragraphs, certain translation phenomena that sentence-level metrics may struggle to handle, like sentence or information reordering, are not well represented in the dataset, limiting our ability to show the benefits of training on paragraph-level ratings.

The contributions of our work include (1) a method for constructing paragraph-level training and meta-evaluation datasets from sentence-level ratings, (2) an experimental study that demonstrates the comparable performance of sentence- and paragraph-level metrics, and (3) an analysis that aims to provide an explanation for our experimental observations.

%% file: 02_terminology.tex
\section{Terminology}
\label{sec:terminology}

Throughout this paper, we use terms like segment, sentence, paragraph, and document to refer to different lengths of text.
To the best of our knowledge, there are no agreed upon definitions for these terms in the MT literature, so here we define how they are used for the rest of the paper.

We refer to the input text to an MT system or evaluation metric as a \emph{segment}, irrespective of its length.
Traditionally, segments in MT have been roughly equivalent to one sentence, although sometimes they can be short phrases or even longer than a single sentence.
Regardless, we use \emph{sentence} to refer to this unit of text since it accurately describes the most common text length that is widely used in MT.

Our work investigates evaluating \emph{paragraphs} of text, which we define to be multi-sentence segments.
We do not require that the paragraphs used in this work obey the traditional definition of a paragraph (i.e., a unit of text separated by a newline character).
We refrain from calling this unit of text a \emph{document}---which we consider to be all of the possible input text---since each document can be broken down into multiple paragraphs and the term paragraph more accurately describes the length of text we use.

%% file: 03_datasets.tex
\section{Paragraph-Level Datasets}
\label{sec:datasets}
The two main sources for training and meta-evaluating MT metrics are the direct assessment (DA) and Multi-dimensional Quality Metrics \citep[MQM; ][]{Lommel14,Freitag21} datasets that the Workshop on Machine Translation (WMT) has collected as part of the yearly metrics shared task \citep{WMT22}.
The DA ratings were done by non-expert raters who assigned a quality score in the range 0-100 to translated sentences.
Because of differences in rater behavior, the DA scores are $z$-normalized per rater.
In MQM, expert raters identify error spans in translated sentences and assign each error a category and severity level, which are used to calculate a score for that error.
A sentence's MQM score is defined as the sum of the errors' scores.

Training and meta-evaluating metrics at the paragraph level requires a collection of translated paragraphs and paragraph-level quality scores.
Luckily, the DA data since 2019 and the MQM data can be considered to be paragraph-level ratings.
The ratings were performed on contiguous blocks of sentences that were translated by the same system (e.g., the first $k$ sentences per document are rated for a system).
Although the scores were collected at the sentence level, the ratings were done in context, meaning the raters had access to the document context for a sentence, so the scores should reflect paragraph- or document-level phenomena like discourse errors.
Therefore, we use the sentence-level DA and MQM data to construct paragraph-level datasets as follows.

\input{figures/number_of_paragraphs_plot}
\input{figures/hypothesis_length_distribution_mqm22}

For each document translated by a system, we run sliding window of size $k$ sentences from the start to the end.
If all $k$ sentences in the window have been rated, those $k$ sentences are concatenated together to become a paragraph instance and the window shifts by $k$.
Otherwise, the sliding window shifts by 1 and the process repeats.
To maintain consistency between the sentence scores within a paragraph, we additionally require that every sentence is scored by the same rater.
Then, we define the paragraph-level scores to be the average DA $z$-score or sum of MQM scores for each sentence in the paragraph.\footnote{
    Summing MQM scores was done to generalize an MQM rating for paragraphs since a sentence's MQM score is the total error weight for that sentence.
    The choice of summing or averaging does not matter for metric meta-evaluation because the correlations are scale invariant.
}
The result is a dataset of rated paragraph translations of $k$ sentences each.

We apply this dataset construction approach to the DA and MQM data for $k = 1, 2, \dots, 10$ sentences per paragraph.
The number of paragraphs is shown in Figure~\ref{fig:number_of_paragraphs} and the distribution of the lengths of the new translated paragraphs is shown in Figure~\ref{fig:hyp_dist_mqm22}.
As $k$ increases, the number of paragraphs decreases because there are fewer candidate paragraphs, while the length of the paragraphs increases, roughly by an expected factor of $k$.

These paragraph-level DA and MQM datasets are used to train and meta-evaluate paragraph-level metrics for the rest of this paper.

%% file: figures/number_of_paragraphs_plot.tex
\begin{figure}
    \centering
    \includegraphics[width=\columnwidth]{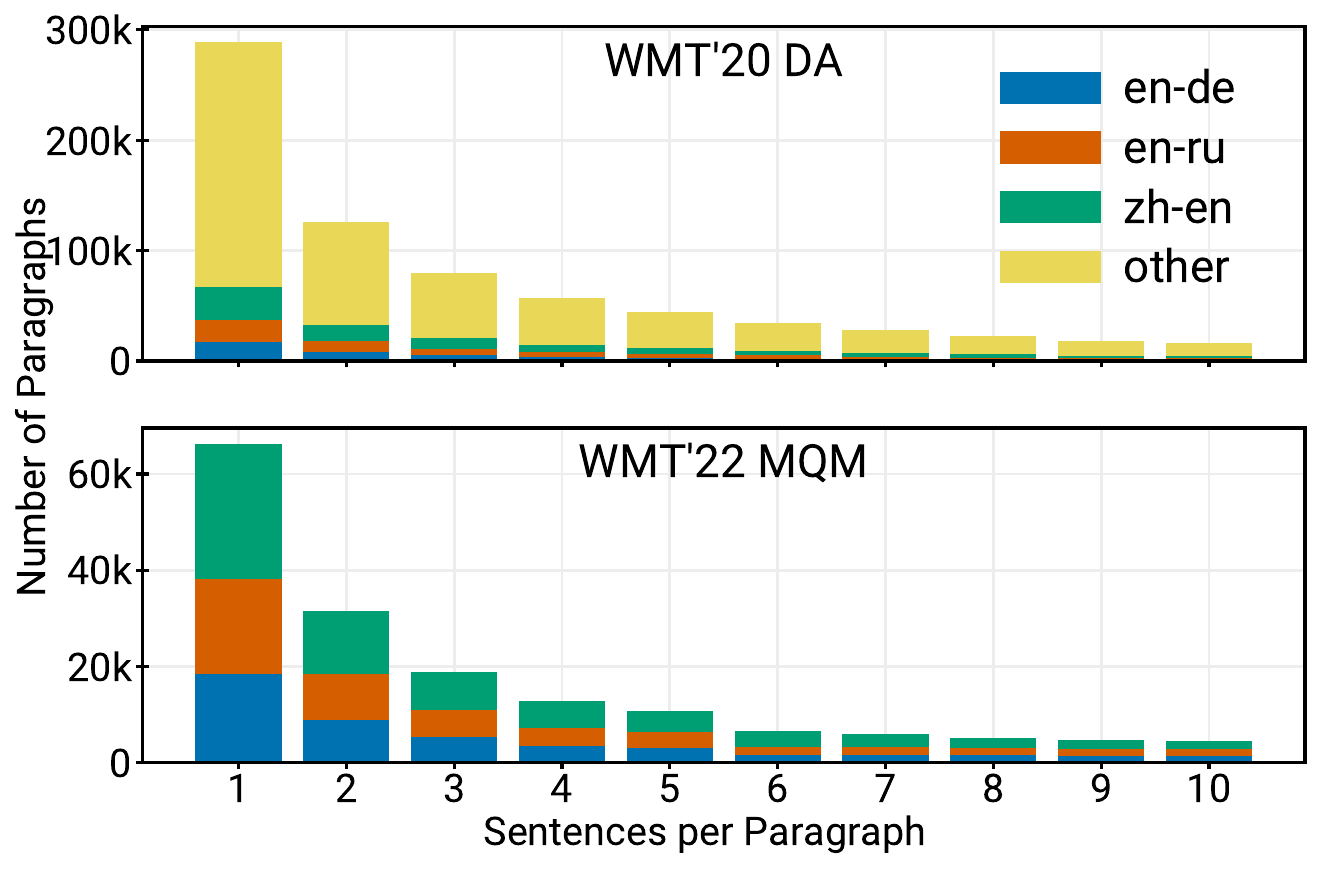}
    \caption{The number of contiguous paragraphs for the given number of sentences per paragraph where each sentence is rated by the same rater.
    Actual values are included in Appendix~\ref{appendix:dataset_stats}.}
    \label{fig:number_of_paragraphs}
\end{figure}

%% file: figures/hypothesis_length_distribution_mqm22.tex
\begin{figure}
    \centering
    \includegraphics[width=\columnwidth]{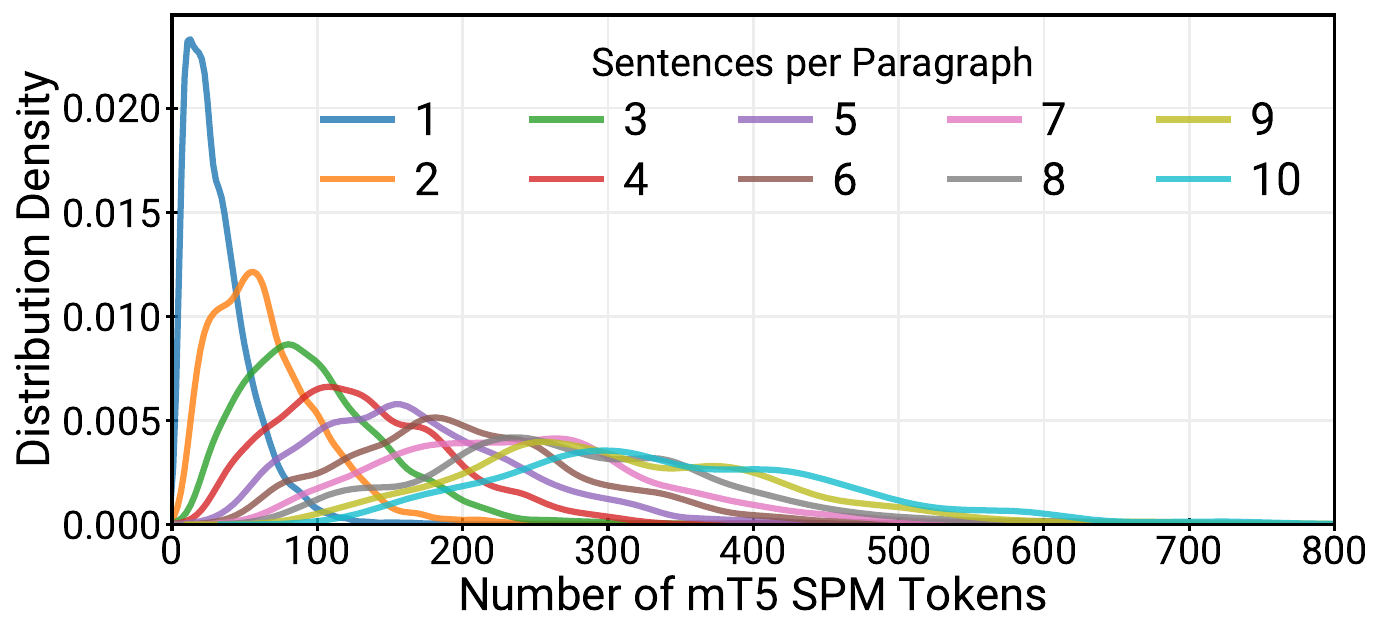}
    \caption{The distribution of SPM token lengths for the WMT'22 MQM dataset for different numbers of sentences per paragraph.
    Additional datasets' distributions are included in Appendix~\ref{appendix:dataset_stats}.
    }
    \label{fig:hyp_dist_mqm22}
\end{figure}

%% file: 04_metrics.tex
\section{Paragraph-Level Metrics}
\label{sec:para_level_metrics}

We explore two different methods for creating paragraph metrics: directly applying sentence-level metrics to paragraphs (\S\ref{sec:seg_to_para}) and training metrics on paragraph-level data (\S\ref{sec:learn_para}).

\subsection{Applying Sentence-Level Metrics on Paragraphs}
\label{sec:seg_to_para}

Although automatic metrics that have been used to evaluate sentence-level MT were not explicitly designed to evaluate paragraphs, they can be repurposed to score paragraphs in different ways.

First, the input paragraph can be treated as if it were one long segment and passed to the metric to calculate a score.
For metrics that use bag-of-$n$-grams representations, like BLEU \citep{papineni-etal-2002-bleu}, there is no input length limitation.
However, some learned metrics, like BLEURT \citep{Sellam20}, have a maximum possible sequence length due to restrictions related to neural network architectures.
Therefore, the length of the input paragraph is restricted in some cases.

Then, if there is assumed to be an alignment between the source, reference, and hypothesis sentences within a paragraph (as is in the case with our datasets), a paragraph score can be calculated by averaging the sentence-level metric's score for each of the $k$ individual sentences.
While this sliding window approach more closely aligns how the metrics are being used to how they were designed, we argue this approach is less than ideal because the 1:1 sentence alignment between the source and hypothesis translations will not always exist.
However, this approach is useful for understanding and analyzing the behavior of metrics when they are used to score full paragraphs directly.

\subsection{Learning Paragraph-Level Metrics}
\label{sec:learn_para}

While sentence-level metrics can be repurposed to score paragraphs, the lengths of the input paragraphs are significantly longer than the lengths of individual sentences (compare $k=1$ to $k>1$ in Figure~\ref{fig:hyp_dist_mqm22}) and there may be cross-sentence dependencies that are not learned by sentence-level metrics.
Therefore, we explore creating a metric specifically for paragraph-level data.

To do so, we train a BLEURT-style regression model on the paragraph-level datasets:
The reference and hypothesis paragraphs are tokenized and concatenated together (separated by a special token), then passed as input to a neural network.
The network is then trained to predict the hypothesis paragraph's ground-truth quality score.
Sections~\ref{sec:metrics} and \ref{sec:implementation_details} contain more information about the model's architecture and implementation details.

It is desirable for the paragraph-level metric to be able to score paragraphs of any length, so we train the metric on paragraphs composed of $k=1,2, \dots, 10$ sentences.
Because the number of paragraph instances decreases significantly as $k$ increases (see Figure~\ref{fig:number_of_paragraphs}), longer paragraphs will rarely be seen during training.
Therefore, we explore two different techniques for weighting training data: one that selects paragraphs uniformly at random and one that performs a stratified sample so the training data is composed of an equal number of paragraphs for each value of $k$.

Next, we describe the experimental setup to evaluate the paragraph-level metrics.

%% file: 05_setup.tex
\section{Experimental Setup}
\label{sec:setup}

\subsection{Datasets}
The paragraph-level datasets used in our experiments are described in Section~\ref{sec:datasets}.
The WMT'19 \citep{WMT19} and '20 \citep{WMT20} paragraph-level DA data is used for training the metrics described in this work, and all metrics are evaluated on the WMT'21 \citep{WMT21} and WMT'22 \citep{WMT22} paragraph-level MQM data.
For both DA and MQM, we use $k=1,2,\dots,10$ sentences per paragraph.
The different paragraph lengths are combined during training but separated for evaluation.

\subsection{Metrics}
\label{sec:metrics}
\paragraph{Paragraph-Level Metrics.}
We train two different paragraph-level metrics, one for each of the different weighting techniques, uniform and stratified sampling (see \S\ref{sec:learn_para}).
We refer to these metrics as \textsc{Para-Unif} and \textsc{Para-Strat}.

Our metric uses the same architecture as the Metric-X WMT'22 metrics shared task submission \citep{WMT22}.
The metric builds on the mT5 encoder-decoder language model \citep{xue-etal-2021-mt5}, which was originally designed to be a sequence-to-sequence language model.
We repurpose the model for our regression task as follows.
The inputs to the encoder are the hypothesis and reference translations separated by a special token, and a single dummy token is passed as the first input to the decoder.
We arbitrarily selected a reserved vocabulary token, then trained the model so that token's output logit in the first decoding step becomes the score for the input hypothesis translation.
This modification of the sequence-to-sequence architecture for regression allows us to utilize all of the pre-trained weights from mT5.

The maximum input sequence length to our metric is 1024 SPM tokens \citep{kudo-richardson-2018-sentencepiece}.
The inputs are truncated during training or inference if the input is larger than 1024.\footnote{
    We experimentally saw no benefit from removing sequences longer than 1024 tokens during training.
}
In the worst case, this happens up to 27\% of the time on the MQM data for 10 sentences per paragraph (see Appendix~\ref{appendix:dataset_stats} for specific statistics.)

\paragraph{Sentence-Level Baseline.}
In addition to the paragraph-level metrics, we train a sentence-level version that is trained on the same DA data but only $k=1$ sentences per paragraph.
This baseline metric can be used to directly compare to the paragraph-level metrics that we train because the model architecture, training procedure, etc., are identical.
The only difference is the training data.
This metric is referred to as \textsc{Sent-Base}.

\paragraph{Other Metrics.}
In addition to the metrics described in this paper, we evaluate BLEU \citep{papineni-etal-2002-bleu}, COMET-22 \citep{Rei20,Rei22}, and PaLM-2 from \citet{fernandes2023devil} as sentence-level metrics applied to paragraphs (i.e., \S\ref{sec:seg_to_para}) and document-level metric BlonDE \citep{jiang-etal-2022-blonde}.
BLEU scores translations using lexical $n$-gram overlap, and COMET-22 is a learned regression metric that first embeds the input hypothesis, reference, and source, combines them to a joint representation, then finally predicts a score.

\input{figures/paragraph_results_zh-en}

The metric from \citet{fernandes2023devil} is based on the PaLM-2 large language model \citep{anil2023palm}.
We evaluate both the zero shot version, in which PaLM-2 is prompted to score a translation on a scale from 0 to 100, and the regression version that finetunes PaLM-2 on MQM ratings to predict a floating point quality score, similar to COMET.
Our analysis includes the Bison variant of PaLM-2.

BlonDE evaluates discourse phenomena in document translations via a set of automatically extracted features.
It was designed to evaluate texts longer than paragraphs, like book chapters, but we compare against it in this work.
BlonDE is available in English only.

We use the SacreBLEU \citep{post-2018-call} implementation of BLEU and the \texttt{Unbabel/wmt22-comet-da} COMET-22 model that was trained on sentence-level WMT DA data from 2017-2020.\footnote{
    Note that the COMET-22 scores we report come from only the reference-based regression model, not the ensemble that was submitted to the WMT'22 metrics shared task.
}

\subsection{Meta-Evaluation Metrics}
The metrics are meta-evaluated using pairwise accuracy at both the system and segment levels.\footnote{
    The segment-level correlation could be referred to as a paragraph-level correlation in this work because the segments we evaluate on are paragraphs.
    However, to be consistent with the evaluation literature, we still use the term segment-level correlation.
}
System-level scores are calculated by averaging a metric's scores over paragraphs.
The system-level pairwise accuracy then computes the proportion of all pairwise system comparison for which the automatic metric and human ground-truth ratings agree on \citep{kocmi-etal-2021-ship}.

We follow \citet{deutsch2023ties} and report segment-level pairwise accuracy using the group-by-item variant of the segment-level correlation in combination with the $\tau$-optimization procedure.
The group-by-item segment-level correlation calculates a pairwise accuracy score between all of the systems' translations for the same input source segment, then averages the accuracy across source segments.
This version of pairwise accuracy gives credit for metrics that correctly predict when two translations have tied ground-truth scores.
Due to the fact that learned regression metrics almost never predict the same score for two non-identical translations, the $\tau$-optimization procedure calibrates the metrics by automatically introducing ties into the metrics' scores.
The segments used in this evaluation are paragraphs.

Both system- and segment-level accuracy (with $\tau$-optimization) are official meta-evaluation metrics for the WMT'23 metrics shared task.\footnote{\url{https://wmt-metrics-task.github.io/}}
Results using Pearson's correlation follow similar trends to the accuracy results and are available in Appendix~\ref{appendix:additional_results}.

\subsection{Implementation Details}
\label{sec:implementation_details}
Our learned metrics are implemented with TensorFlow \citep{tensorflow2015-whitepaper} in the T5X library \citep{roberts2022t5x}.
They are initialized with the XXL version of mT5, which contains 13B parameters.
It is trained for a maximum of 20k steps and a batch size of 128 using Adafactor \citep{shazeer2018adafactor} on 64 v3 TPUs.
Checkpoint selection was done by selecting the step that has the highest average segment-level pairwise accuracy across language pairs and all values of $k$ sentences per paragraph after applying calibrating via $\tau$-optimization.
In general, we observed the specific checkpoint selection strategy was not too important.

%% file: figures/paragraph_results_zh-en.tex
\begin{figure*}[t]
    \centering
    \includegraphics[width=\textwidth]{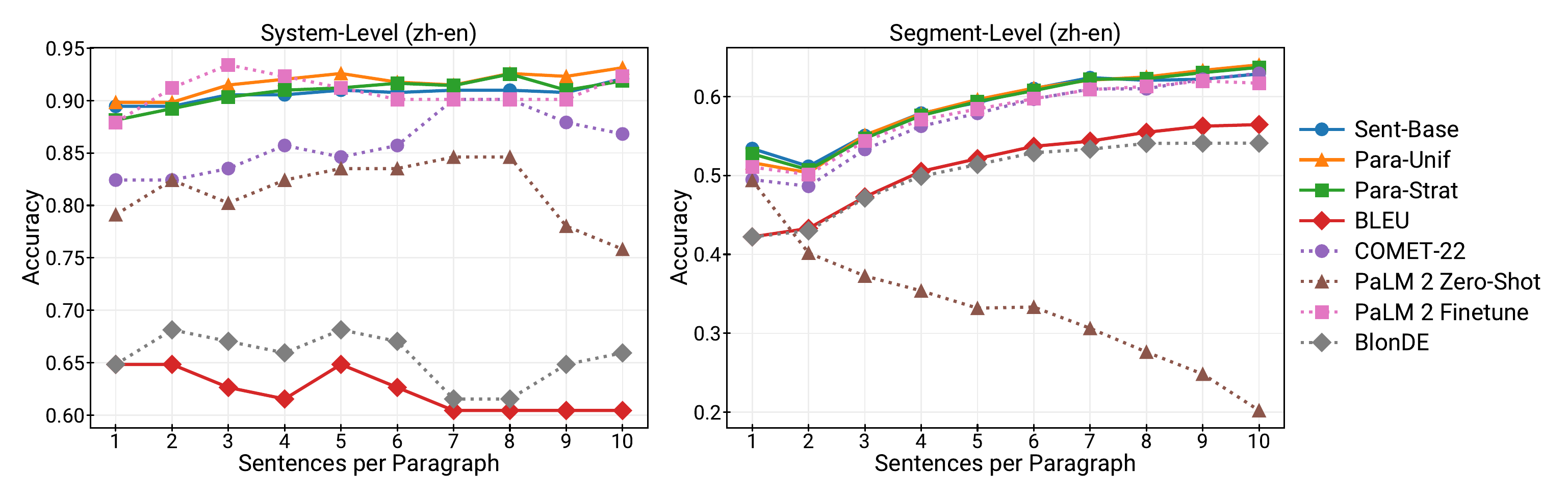}
    \caption{As the number of sentences per paragraph increases, the accuracy scores of the metrics appears to either not decrease (system-level, left) or increase (segment-level, right).
    This suggests that accurately scoring a paragraph is an easier task than an individual sentence, even for metrics that are not trained on paragraph-level examples.
    The results of metrics trained in this work presented here are an average of 5 different runs.
    Results for other language pairs follow the same trend and are included in Appendix~\ref{appendix:additional_results}.
    }
    \label{fig:paragraph_results_zh_en}
\end{figure*}

%% file: 06_results.tex
\section{Results}
\label{sec:results}

First, we directly evaluate how well metrics perform when used to directly score paragraphs (\S\ref{sec:para_eval}), then we further examine the behavior of different paragraph-level metrics by analyzing their performances with the context of their sentence-level counterparts (\S\ref{sec:seg_comparison}).

\subsection{Paragraph-Level Evaluation}
\label{sec:para_eval}

Figure~\ref{fig:paragraph_results_zh_en} plots the system- and segment-level correlation results for different numbers of $k$ sentences per paragraph.
Each metric is used to directly score a full paragraph even if the metric was not designed to do so (e.g., \textsc{Sent-Base} or COMET-22).
There are several interesting observations.

\paragraph{Paragraph-Level Performance.}
First, as the length of the paragraphs increases, the system-level correlations remain relatively steady or increase and the segment-level correlations clearly improve for all metrics, except for PaLM-2 zero-shot.
This is evidence that scoring paragraphs is an easier task than scoring individual sentences, a result that is counterintuitive;
scoring more text should seemingly be a harder task.
We hypothesize this result is explained by the fact that some noise in the human and metric scores is averaged away, leaving more reliable signals as the paragraphs get longer.
If the metric scores are unbiased estimators, their agreement with human rating should then increase.

PaLM-2 zero-shot is an outlier in this case because it predicts a large number of ties between translations.
Prompting large language models for MT evaluation is known to result in the model predicting a small number of unique scores, resulting in many ties \citep{gemba,fernandes2023devil}.
As the length of the paragraph increases, the number of MQM ties decreases.
Since pairwise accuracy penalizes incorrect tie predictions, the zero shot model has worse performance on longer texts.
See Figure~\ref{fig:percent_ties_ende} for a visualization of the number of ties in the PaLM-2 output and MQM scores.

\input{figures/percent_ties_en-de}

\paragraph{Sentence vs. Paragraph Level.}

Then, there appears to be little evidence that training on paragraph-level examples results in better correlations to human ratings on paragraph-level test data.
For instance, increasing the weight of the paragraph-level data during training does not help compared to uniformly sampling data (compare \textsc{Para-Strat} to \textsc{Para-Unif}).
Further, the baseline metric \textsc{Sent-Base} that shares the same architecture as our paragraph-level metrics but is only trained on sentence-level data ($k=1$) performs just as well as the paragraph-level metrics.
This observation is additionally supported by COMET-22's results.
The difference between the metrics we train versus COMET is relatively constant for all values of $k$, demonstrating that COMET is not systematically worse on longer inputs.

The generalization of sentence-level metrics on paragraph-level data is rather surprising.
The length of the inputs for scoring paragraphs is up to 10x longer than those for scoring sentences (see Table~\ref{tab:percentiles}).
Even though the length of the test data is out-of-distribution with respect to the training data, the sentence-level metrics predict reliable scores on the paragraph-level data.
Next, we further analyze the sentence-level metrics to better understand their scores.

\input{figures/hypothesis_length_percentiles}

\subsection{Understanding Sentence-Level Metrics}
\label{sec:seg_comparison}

To further analyze the performance of the sentence-level metrics on paragraph-level data, we compare the two versions of applying a sentence-level metric to paragraphs discussed in \S\ref{sec:seg_to_para}.
One version directly scores a full paragraph (thus, making no assumption about an alignment between the hypothesis and reference), whereas the other averages the scores of evaluating the individual $k$ hypothesis sentences against the corresponding reference sentence (thus, assuming a sentence-level alignment exists).

Figure~\ref{fig:alignment_plot_en_de} shows that for two sentence-level metrics, the baseline trained in this work and BLEU, the performance of the two paragraph scoring variants is very similar.
Then, Figure~\ref{fig:self_correlation} shows that the Pearson correlation between the scores for those two variants is very high ($\geq 0.85$).

\input{figures/alignment_comparison_en-de}

Together, these results point to the fact that there is little difference between these two methods.
Directly scoring a paragraph or scoring individual sentences yield both similar scores and similar agreement to human ratings.
The sentence-level metrics appear to be scoring full paragraphs in a desirable way---by calculating some average score across sentences.

This result is not obvious.
As the length of the input increases, the bag-of-$n$-grams representation used by lexical matching metrics like BLEU have an increased potential for erroneous matches between the hypothesis and reference sentences, which could result in misleading scores.
Learned metrics, like the ones trained in this work, have not been trained on a significant amount of very long data, so it is not clear that the scoring functions they learn would generalize well to longer inputs.
Despite this, the sentence-level metrics appear to predict high-quality scores for paragraphs.

Next, we propose a hypothesis for why this is the case and why training on paragraph-level data does not appear to result in a better metric.

\input{figures/self_correlation}

%% file: figures/percent_ties_en-de.tex
\begin{figure}
    \centering
    \includegraphics[width=\columnwidth]{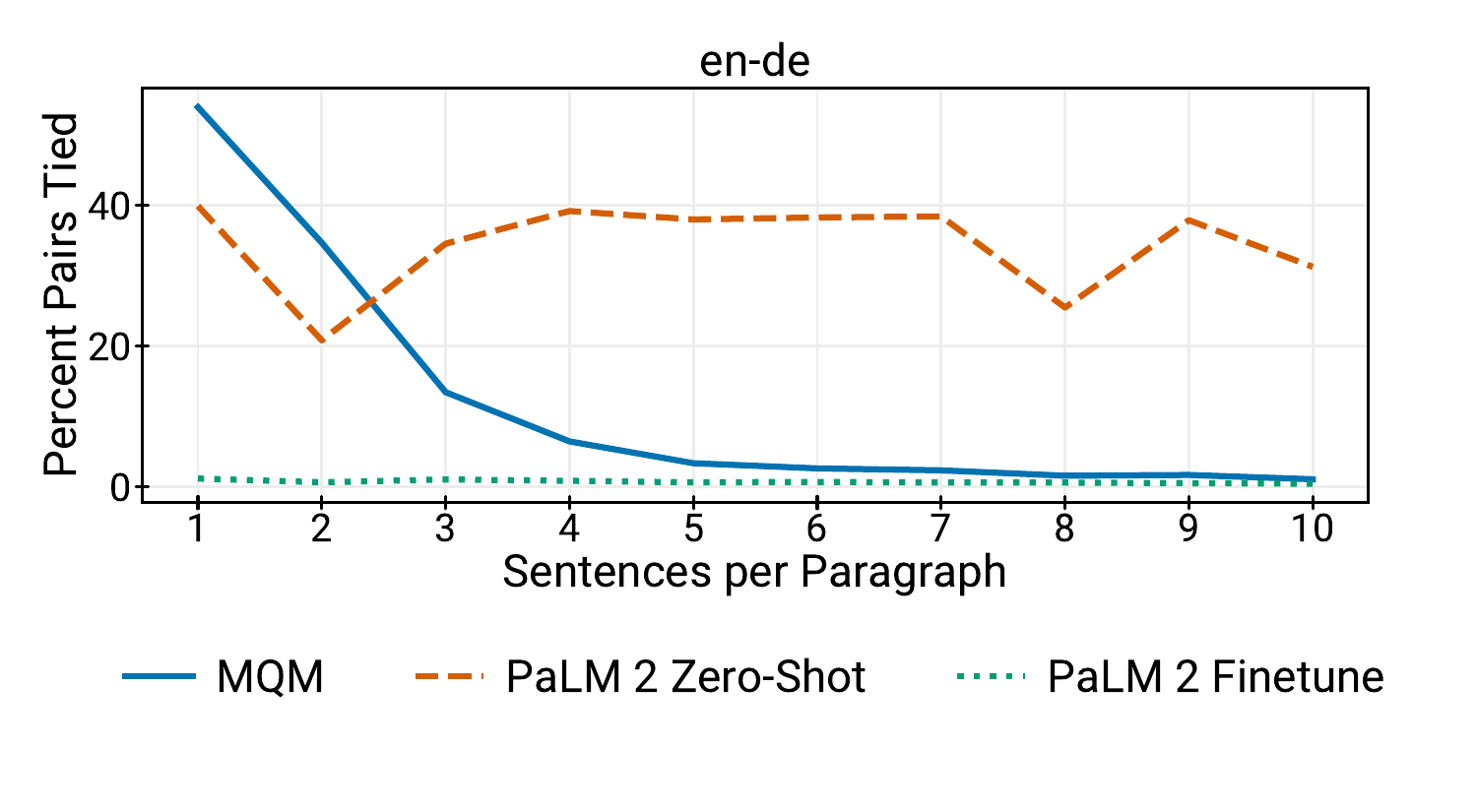}
    \caption{There are fewer MQM ties as the number of sentences per paragraph increases.
    The finetuned PaLM-2 model outputs a very small number of ties, whereas the zero-shot model consistently predicts a large number of ties.
    Since the pairwise accuracy meta-evaluation metric penalizes metrics for incorrect tie predictions, the zero-shot model will have worse performance as the inputs get longer.
    }
    \label{fig:percent_ties_ende}
\end{figure}

%% file: figures/hypothesis_length_percentiles.tex
\begin{table}[t]
    \centering
    \begin{adjustbox}{width=\columnwidth}
\begin{tabular}{lrrrrrrr}
\toprule
     \multirow{2}{*}{\bf Dataset} &      \multicolumn{3}{c}{\bf 1 Sent. per Para.} &   &   \multicolumn{3}{c}{\bf 10 Sent. per Para.}\\
     \cmidrule{2-4} \cmidrule{6-8}
     & 25th & 50th & 75th & & 25th & 50th & 75th \\
\midrule
  WMT'19 DA &    20 &    31 &    47 & &   300 &    362 &    431 \\
  WMT'20 DA &    24 &    38 &    58 & &   318 &    410 &    524 \\
 WMT'21 MQM &    28 &    41 &    57 & &   370 &    433 &    516 \\
 WMT'22 MQM &    15 &    27 &    43 & &   265 &    333 &    426 \\
\bottomrule
\end{tabular}

    \end{adjustbox}
    \caption{The SPM token lengths for the given percentiles are in general around 10 times larger with 10 sentences per paragraph compared to a single sentence.
    Visualizations of the distributions for every paragraph length can be found in Appendix~\ref{appendix:dataset_stats}.}
    \label{tab:percentiles}
\end{table}

%% file: figures/alignment_comparison_en-de.tex
\begin{figure}
    \centering
    \includegraphics[width=\columnwidth]{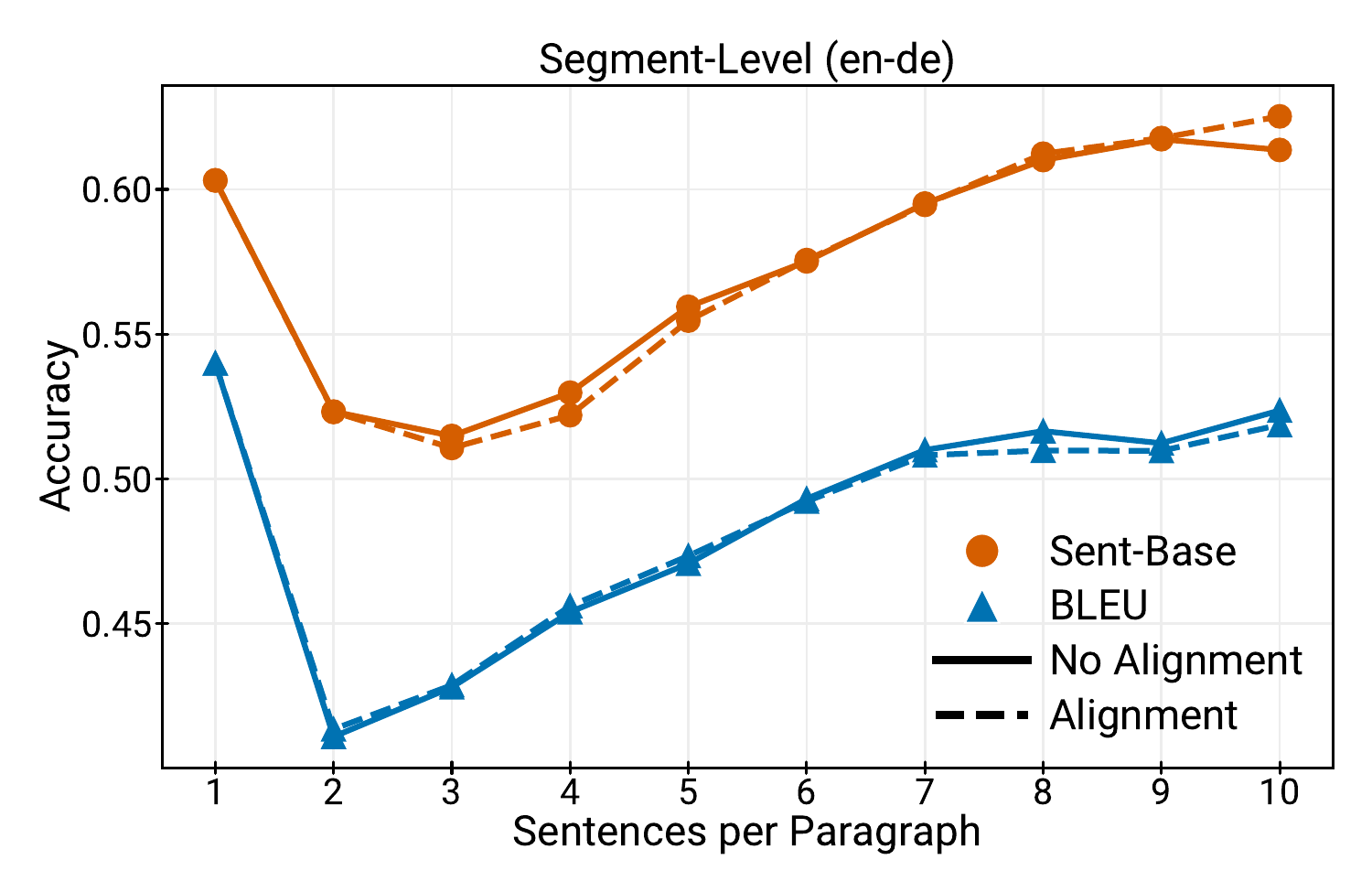}
    \caption{
        Metrics that score a paragraph directly (solid line) versus those that assume an alignment between the reference and hypothesis and calculate a score by averaging across the $k$ sentence-level values (dashed line) perform very similarly.
    }
    \label{fig:alignment_plot_en_de}
\end{figure}

%% file: figures/self_correlation.tex
\begin{figure}
    \centering
    \includegraphics[width=\columnwidth]{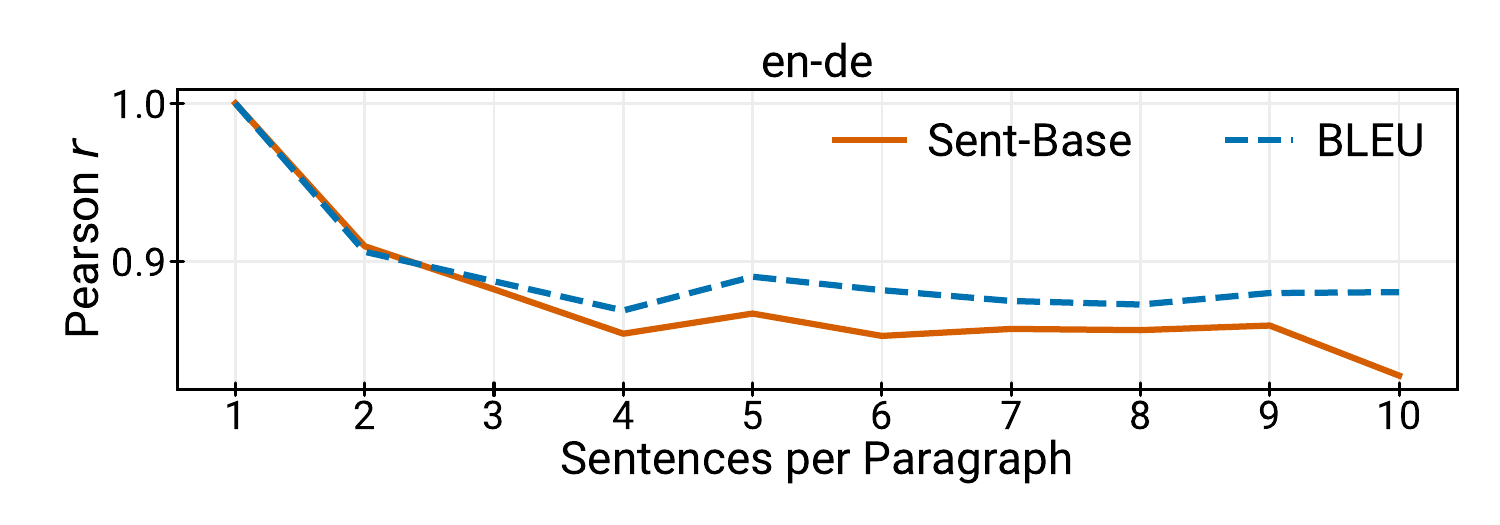}
    \caption{The plot shows the Pearson correlation on en-de between directly predicting a score for a paragraph of $k$ sentences and calculating a paragraph score by averaging over $k$ sentence-level scores.
    The correlations are quite high, demonstrating that the both methods result in very similar scores.
    }
    \label{fig:self_correlation}
\end{figure}

%% file: 07_discussion.tex
\section{Discussion}
\label{sec:discussion}

In theory, training on paragraph-level data should have advantages compared to training on sentence-level data.
The metric (1) should be able to handle longer input sequences, (2) it should be able to capture long range dependencies, and (3) it should be able to model different paragraph-level phenomena like information or sentence reordering.
However, we were not able to demonstrate these advantages in practice, and we theorize why as follows.

First, the analysis in \S\ref{sec:seg_comparison} shows that sentence-level metrics generalize well to significantly longer input, so advantage (1) may not be so relevant.
We hypothesize that the scoring function learned by sentence-level metrics like \textsc{Sent-Base} or COMET could score a token in the hypothesis based on some alignment to the reference using its relative position in the translation.
This function would be agnostic with respect to the global positioning, and thus the scoring function would generalize well to longer inputs.
If this were true, training on paragraph-level data would not be necessary to obtain good performance on long sequences.

Second, evaluating translation quality seems to be a very ``local'' problem in the sense that modeling long range dependencies is not frequently necessary for evaluation.
Often, the reference phrase that aligns to a hypothesis phrase has enough information to accurately evaluate the hypothesis.
If it does not, the information is likely nearby, not several sentences away (see Figure~\ref{fig:maria}).
Although the sentence-level metrics were not trained on multiple sentences, we suspect they are able to capture nearby dependencies across sentences when evaluating paragraphs.
In theory, a paragraph-level metric would have the ability to model long range dependencies since it could observe them during training.
However, if they are infrequent, advantage (2) over sentence-level metrics may be small.

\input{figures/maria}

Finally, the ability for our learned paragraph metrics to capture phenomena like sentence reordering is limited by our dataset construction method.
Since the paragraphs in our training and test sets come from MT systems that translated one sentence at a time, there are no phenomena like sentence reordering present in the datasets.
Therefore, the paragraph-level metric cannot learn to model such cases, and the metrics are never evaluated on them either.
Thus, the limitations of the dataset mean that we cannot demonstrate advantage (3).

We believe that paragraph-level metrics are necessary for evaluating true paragraph translations, where MT systems can be more creative with how a full paragraph is translated, rather than paragraph translations that are created by translating individual sentences.
We hypothesize that sentence-level metrics will not generalize well when there is no sentence alignment or there is significant information reordering.
To accurately evaluate actual paragraph translations, metrics need to be trained on similar data.
Future work should invest in collecting human ratings for paragraph-level translations so that new metrics can be trained and evaluated.

%% file: figures/maria.tex
\begin{figure}[t]
    \begin{mdframed}
    \small
        \textbf{Source Context:} Maria said no.
        
        \textbf{Source:} She did not slap the green witch.
        \\ \\
        \textbf{Reference Context:} Maria dijo no.
        
        \textbf{Reference:} No le di\'{o} una bofetada a la bruja verde.
        \\ \\
        \textbf{Hypothesis:} Ella({\color{OliveGreen}\cmark})/\'{E}l({\color{red}\xmark}) no le di\'{o} una bofetada a la bruja verde.
    \end{mdframed}
    \caption{An English-to-Spanish translation example where the reference translation does not have enough information to correctly evaluate the hypothesis.
    Gender in Spanish is marked on pronouns, and Spanish is a pro-drop language, which means the pronoun can be omitted if the context is clear.
    In this example, the pronoun is dropped from the reference, so determining whether the pronoun used in the hypothesis requires taking into account the previous reference sentence.
    We suspect such examples are not frequent, and if they do exist, the information required to resolve the ambiguity is relatively local to the reference sentence.
    }
    \label{fig:maria}
\end{figure}

%% file: 08_related_work.tex
\section{Related Work}

The vast majority of research on MT evaluation has worked at the sentence level \citep{papineni-etal-2002-bleu,banerjee-lavie-2005-meteor,snover-etal-2006-study,popovic-2015-chrf,popovic-2017-chrf,lo-2019-yisi,Sellam20,Rei20,Rei22,thompson-post-2020-automatic,wan-etal-2022-unite}, although there has been recent interest in moving beyond sentence-level evaluation.
\citet{vernikos-etal-2022-embarrassingly} propose a method to incorporate document-level context into a sentence-level metric by using the additional context when computing the representations for the hypothesis and reference sentences.
Although they use document context in their metric, it is still scores single sentences at a time, in contrast to the paragraph-level metrics in our work that predict a score for entire paragraphs at once.
Then \citet{jiang-etal-2022-blonde} propose a document-level metric called BlonDE that targets evaluating discourse phenomena as opposed to overall translation quality (i.e., they do not model translation accuracy errors).
To the best of our knowledge, ours is the first study aimed at training a learned metric that directly scores entire paragraphs.

Other studies that have evaluated sentence-level metrics beyond the sentence-level have done so in the literary domain.
\citet{thai-etal-2022-exploring} show that automatic metrics prefer MT output over human translations, and  \citet{karpinska2023large} show that metrics prefer actual translations of paragraphs over sentence-by-sentence translations.
Our work is complementary to theirs as we focus on the news domain, train metrics on paragraph-level data, and evaluate on a much larger set of human ratings.
It is not clear whether conclusions reached about metrics in the news domain will apply to the literary domain or vice versa.

Some researchers have developed challenge sets that can be used to probe how well metrics capture discourse phenomena that appear when translating more than one sentence at a time \citep{bawden-etal-2018-evaluating,muller-etal-2018-large,lopes-etal-2020-document}.
However, these challenge sets can be trivial for reference-based metrics because the reference often resolves the ambiguity in the translation.
To the best of our knowledge, a challenge set that forces reference-based metrics to use context outside of a single reference sentence during evaluation (see Figure~\ref{fig:maria}) does not exist.

Research on generating translations of text longer than single sentences directly use sentence-level metrics to score translations \citep{tiedemann-scherrer-2017-neural,miculicich-etal-2018-document,ma-etal-2020-simple,wu-etal-2023-document,post2023escaping}.
Our work can be viewed as a justification for doing so.

%% file: 09_conclusion.tex
\section{Conclusion}
In this work, we proposed a method for constructing paragraph-level datasets for training and meta-evaluating MT evaluation metrics from sentence-level data.
Our experimental results showed that metrics trained on paragraph-level data do not necessarily out-perform those trained on sentence-level data, potentially due to the fact that sentence-level metrics seem to generalize well to longer inputs and limitations of our paragraph-level datasets.
Future work should invest in collecting human judgments for paragraph translations generated by MT systems that directly translate full paragraphs instead of translating one sentence at a time.
Such a dataset would be more likely to contain phenomena that do not exist at the sentence level, which we hypothesize would be more likely to require metrics designed to work at the paragraph level.

\section*{Acknowledgments}
The authors would like to thanks George Foster and Macduff Hughes for their valuable feedback on this work.

%% file: 10_limitations.tex
\section*{Limitations}
There are a couple of limitations related to our dataset construction approach that are worth enumerating.

As discussed in Section~\ref{sec:discussion}, our ability to evaluate metrics' performances on all types of paragraph-level translations is limited by our dataset construction method.
Our translated paragraphs are generated by MT systems that translate one sentence at time, which results in sentence aligned data.
Therefore, we are unable to evaluate metrics on true paragraph-level translations that might have sentence or information reordering.

Then, the WMT data no longer contains information about the white space between the original source sentences.
Therefore, the DA and MQM paragraph-level datasets do not contain the paragraph breaks that were in the original document.
Each of the $k$ sentences is concatenated together and separated by a space in our work, so it is very likely that the artificially constructed paragraphs do not perfectly resemble real paragraphs.

%% file: appendix/dataset_statistics.tex
\section{Dataset Statistics}
\label{appendix:dataset_stats}

The exact number of paragraph-level instances by WMT year and language pair that we generaetd from our dataset construction procedure (see \S\ref{sec:datasets}) can be found in Table~\ref{tab:da_stats} for DA and Table~\ref{tab:mqm_stats} for MQM.
Figure~\ref{fig:hyp_dist_all} visualizes the distribution of the lengths of the hypotheses in the paragraph-level datasets based on mT5 SPM tokens.
Then, Table~\ref{tab:too_long} contains the number of paragraph examples that are too long to fit into the 1024 SPM maximum context length that is used by the metrics trained in this work.

\input{figures/da_paragraph_counts}
\input{figures/mqm_paragraph_counts}
\input{figures/hypothesis_distribution_all}
\input{figures/too_long_inputs}

%% file: figures/da_paragraph_counts.tex
\begin{table*}[t]
    \centering

\begin{tabular}{rlrrrrrrrrrr}
\toprule
 \multirow{2}{*}{\bf Year} &     \multirow{2}{*}{\bf LP} &      \multicolumn{10}{c}{\bf Sentences per Paragraph} \\
 \cmidrule{3-12}
  & &   
  \multicolumn{1}{c}{\bf 1} &
 \multicolumn{1}{c}{\bf 2} &
 \multicolumn{1}{c}{\bf 3} &
 \multicolumn{1}{c}{\bf 4} &
 \multicolumn{1}{c}{\bf 5} &
 \multicolumn{1}{c}{\bf 6} &
 \multicolumn{1}{c}{\bf 7} &
 \multicolumn{1}{c}{\bf 8} &
 \multicolumn{1}{c}{\bf 9} &
 \multicolumn{1}{c}{\bf 10}\\
\midrule
 2019 &  de-cs &  16900 &   1032 &     95 &     12 &     1 &     0 &     0 &     0 &     0 &     0 \\
 2019 &  de-en &  34756 &  16754 &  10896 &   7735 &  5976 &  4660 &  3947 &  3147 &  2730 &  2345 \\
 2019 &  de-fr &   6700 &    173 &      5 &      0 &     0 &     0 &     0 &     0 &     0 &     0 \\
 2019 &  en-cs &  27445 &  13241 &   8710 &   6152 &  4834 &  3865 &  3215 &  2607 &  2371 &  1967 \\
 2019 &  en-de &  45131 &  21777 &  14311 &  10124 &  7932 &  6363 &  5274 &  4285 &  3906 &  3232 \\
 2019 &  en-fi &  20618 &   9937 &   6557 &   4611 &  3628 &  2910 &  2419 &  1945 &  1799 &  1482 \\
 2019 &  en-gu &  10151 &   4890 &   3229 &   2267 &  1774 &  1423 &  1221 &   964 &   884 &   722 \\
 2019 &  en-kk &  12922 &   6221 &   4115 &   2888 &  2253 &  1813 &  1562 &  1223 &  1112 &   910 \\
 2019 &  en-lt &  13217 &   6363 &   4219 &   2963 &  2319 &  1863 &  1603 &  1257 &  1137 &   944 \\
 2019 &  en-ru &  22600 &  10902 &   7180 &   5069 &  3974 &  3185 &  2650 &  2137 &  1966 &  1633 \\
 2019 &  en-zh &  26530 &  12810 &   8434 &   5944 &  4673 &  3758 &  3102 &  2520 &  2308 &  1904 \\
 2019 &  fi-en &  20286 &    362 &     21 &      2 &     0 &     0 &     0 &     0 &     0 &     0 \\
 2019 &  fr-de &   4000 &     87 &      3 &      0 &     0 &     0 &     0 &     0 &     0 &     0 \\
 2019 &  gu-en &  14860 &    550 &     40 &      2 &     0 &     0 &     0 &     0 &     0 &     0 \\
 2019 &  kk-en &  15763 &    705 &     77 &     10 &     0 &     0 &     0 &     0 &     0 &     0 \\
 2019 &  lt-en &  16046 &    489 &     32 &      2 &     0 &     0 &     0 &     0 &     0 &     0 \\
 2019 &  ru-en &  24247 &    785 &     83 &     10 &     1 &     0 &     0 &     0 &     0 &     0 \\
 2019 &  zh-en &  50722 &  15164 &   9347 &   6774 &  5030 &  4087 &  3312 &  2714 &  2226 &  1797 \\
 \midrule
 2020 &  cs-en &   9381 &   4322 &   2628 &   1797 &  1323 &   940 &   685 &   404 &   241 &   138 \\
 2020 &  de-en &  12541 &   5825 &   3451 &   2422 &  1808 &  1220 &   927 &   652 &   507 &   378 \\
 2020 &  en-cs &  34180 &  16371 &  10324 &   7358 &  5591 &  4501 &  3474 &  2749 &  2270 &  2035 \\
 2020 &  en-de &  17393 &   8337 &   5253 &   3723 &  2859 &  2283 &  1729 &  1362 &  1138 &  1033 \\
 2020 &  en-iu &   6145 &   3028 &   1990 &   1479 &  1152 &   937 &   801 &   693 &   600 &   538 \\
 2020 &  en-ja &  21999 &  10672 &   6769 &   5036 &  3907 &  3093 &  2513 &  2109 &  1812 &  1635 \\
 2020 &  en-pl &  18342 &   8891 &   5636 &   4192 &  3266 &  2569 &  2089 &  1756 &  1514 &  1377 \\
 2020 &  en-ru &  19543 &   9494 &   6058 &   4433 &  3477 &  2750 &  2279 &  1847 &  1602 &  1468 \\
 2020 &  en-ta &   9175 &   4439 &   2825 &   2100 &  1634 &  1301 &  1035 &   875 &   746 &   680 \\
 2020 &  en-zh &  41965 &  20069 &  12656 &   9034 &  6843 &  5510 &  4260 &  3371 &  2782 &  2483 \\
 2020 &  iu-en &  12172 &     75 &      0 &      0 &     0 &     0 &     0 &     0 &     0 &     0 \\
 2020 &  ja-en &   9879 &   4710 &   3047 &   2103 &  1715 &  1321 &  1053 &   845 &   759 &   639 \\
 2020 &  km-en &   6951 &     72 &      0 &      0 &     0 &     0 &     0 &     0 &     0 &     0 \\
 2020 &  pl-en &  12435 &   6048 &   3871 &   2857 &  2184 &  1708 &  1445 &  1265 &  1030 &   844 \\
 2020 &  ps-en &   7138 &    110 &      2 &      0 &     0 &     0 &     0 &     0 &     0 &     0 \\
 2020 &  ru-en &  11244 &   5369 &   3408 &   2405 &  1832 &  1488 &  1179 &   952 &   785 &   604 \\
 2020 &  ta-en &   7842 &   3762 &   2406 &   1723 &  1322 &  1065 &   847 &   694 &   572 &   473 \\
 2020 &  zh-en &  30325 &  14567 &   9253 &   6674 &  5106 &  4078 &  3374 &  2811 &  2223 &  1824 \\
\bottomrule
\end{tabular}

    \caption{The number of paragraphs with the given number of sentences per paragraph from the direct assessment data from WMT'19 and WMT'20.
    Each paragraph is required to a contiguous block of sentences that are rated by the same rater.
    }
    \label{tab:da_stats}
\end{table*}

%% file: figures/mqm_paragraph_counts.tex
\begin{table*}[t]
    \centering

\begin{tabular}{ccrrrrrrrrrr}
\toprule
 \multirow{2}{*}{\bf Dataset} &     \multirow{2}{*}{\bf LP} &      \multicolumn{10}{c}{\bf Sentences per Paragraph} \\
 \cmidrule{3-12}
  & &
  \multicolumn{1}{c}{\bf 1} &
 \multicolumn{1}{c}{\bf 2} &
 \multicolumn{1}{c}{\bf 3} &
 \multicolumn{1}{c}{\bf 4} &
 \multicolumn{1}{c}{\bf 5} &
 \multicolumn{1}{c}{\bf 6} &
 \multicolumn{1}{c}{\bf 7} &
 \multicolumn{1}{c}{\bf 8} &
 \multicolumn{1}{c}{\bf 9} &
 \multicolumn{1}{c}{\bf 10}\\
\midrule
 WMT'21 &  en-de &   7905 &   3825 &  2460 &  1800 &  1395 &  1140 &   870 &   765 &   660 &   585 \\
 WMT'21 &  en-ru &   7905 &   3825 &  2460 &  1800 &  1395 &  1140 &   870 &   765 &   660 &   585 \\
 WMT'21 &  zh-en &   9058 &   4340 &  2814 &  1974 &  1596 &  1190 &   994 &   770 &   658 &   644 \\
 \midrule
      WMT'22 &  en-de &  18410 &   8932 &  5236 &  3486 &  3080 &  1610 &  1568 &  1470 &  1372 &  1330 \\
      WMT'22 &  en-ru &  19725 &   9570 &  5610 &  3735 &  3300 &  1725 &  1680 &  1575 &  1470 &  1425 \\
      WMT'22 &  zh-en &  28110 &  13005 &  7935 &  5655 &  4245 &  3285 &  2670 &  2160 &  1935 &  1710 \\
\bottomrule
\end{tabular}

    \caption{The number of paragraphs with the given number of sentences per paragraph from the MQM data from WMT'21 and WMT'22.
    Each paragraph is required to a contiguous block of sentences that are rated by the same rater.
    }
    \label{tab:mqm_stats}
\end{table*}

%% file: figures/hypothesis_distribution_all.tex
\begin{figure}
    \centering
    \includegraphics[width=\columnwidth]{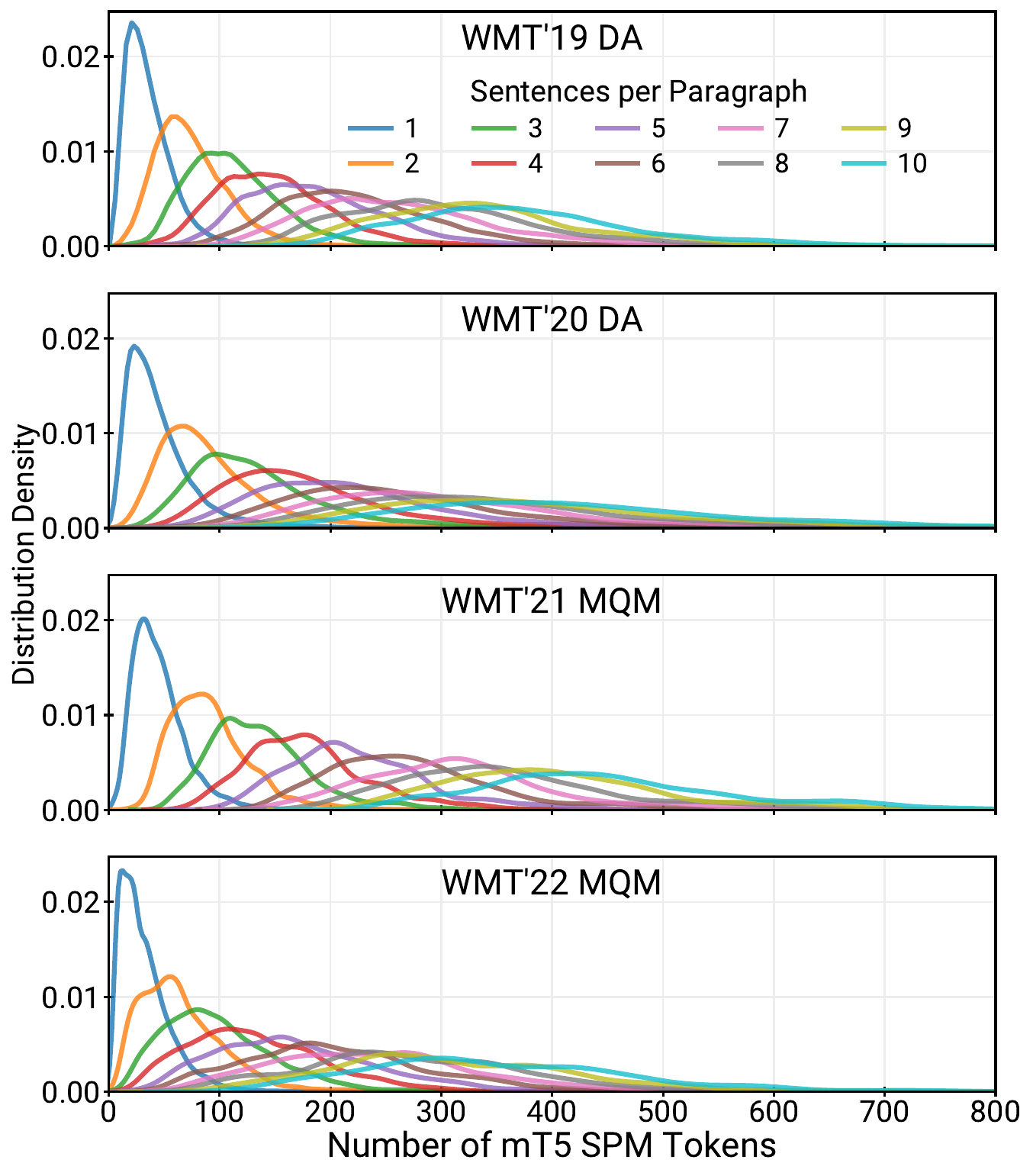}
    \caption{The distribution of the length of the hypothesis translations for the direct assessment (DA) and MQM datasets for a given number of sentences per paragraph.}
    \label{fig:hyp_dist_all}
\end{figure}

%% file: figures/too_long_inputs.tex
\begin{table*}[t]
    \centering
    \begin{adjustbox}{width=\textwidth}

\begin{tabular}{lrrrrrrrrrr}
    \toprule
 \multirow{2}{*}{\bf Dataset} &      \multicolumn{10}{c}{\bf Sentences per Paragraph} \\
 \cmidrule{2-11}
 & 
 \multicolumn{1}{c}{\bf 1} &
 \multicolumn{1}{c}{\bf 2} &
 \multicolumn{1}{c}{\bf 3} &
 \multicolumn{1}{c}{\bf 4} &
 \multicolumn{1}{c}{\bf 5} &
 \multicolumn{1}{c}{\bf 6} &
 \multicolumn{1}{c}{\bf 7} &
 \multicolumn{1}{c}{\bf 8} &
 \multicolumn{1}{c}{\bf 9} &
 \multicolumn{1}{c}{\bf 10}
    \\ 
  \midrule
  WMT'19 DA &  2 (0\%) &    3 (0\%) &    4 (0\%) &    15 (0\%) &    48 (0\%) &   196 (1\%) &    440 (2\%) &    702 (3\%) &   1349 (7\%) &  1944 (11\%) \\
  WMT'20 DA &  4 (0\%) &  179 (0\%) &  667 (1\%) &  1148 (2\%) &  1598 (4\%) &  2222 (6\%) &  2879 (10\%) &  3389 (15\%) &  4041 (22\%) &  4688 (29\%) \\
 WMT'21 MQM &  0 (0\%) &    0 (0\%) &    0 (0\%) &     0 (0\%) &     3 (0\%) &    23 (1\%) &    103 (4\%) &   245 (11\%) &   295 (15\%) &   488 (27\%) \\
 WMT'22 MQM &  0 (0\%) &    0 (0\%) &    6 (0\%) &    11 (0\%) &    56 (1\%) &    74 (1\%) &    110 (2\%) &    202 (4\%) &    266 (6\%) &   450 (10\%) \\
\bottomrule
\end{tabular}

\end{adjustbox}
    \caption{The number (and percent) of paragraphs for which the number of SPM tokens in the reference and hypothesis combined is larger than the maximum allowable input length by our metric, 1024.
    If the input is too long, it is truncated.
    }
    \label{tab:too_long}
\end{table*}

%% file: appendix/additional_results.tex
\section{Additional Results}
\label{appendix:additional_results}

Figure~\ref{fig:paragraph_results_ende_enru} contains the system- and segment-level accuracy correlations on the en-de and en-ru language pairs from WMT'22 MQM that were not presented in the main body of the paper.
Figure~\ref{fig:paragraph_pearson} contains the correlations for all 3 language pairs but uses Pearson correlation instead of pairwise accuracy.

Figure~\ref{fig:self_correlation_enru_zhen} shows the correlation between the two ways to apply a segment-level metric to paragraph-level data, directly scoring the paragraph or averaging the $k$ segment scores, on the en-ru and zh-en WMT'22 MQM dataset.

\input{figures/paragraph_results_appendix}
\input{figures/paragraph_results_pearson}

\input{figures/self_correlation_appendix}

%% file: figures/paragraph_results_appendix.tex
\begin{figure*}
    \centering
    \includegraphics[width=\textwidth]{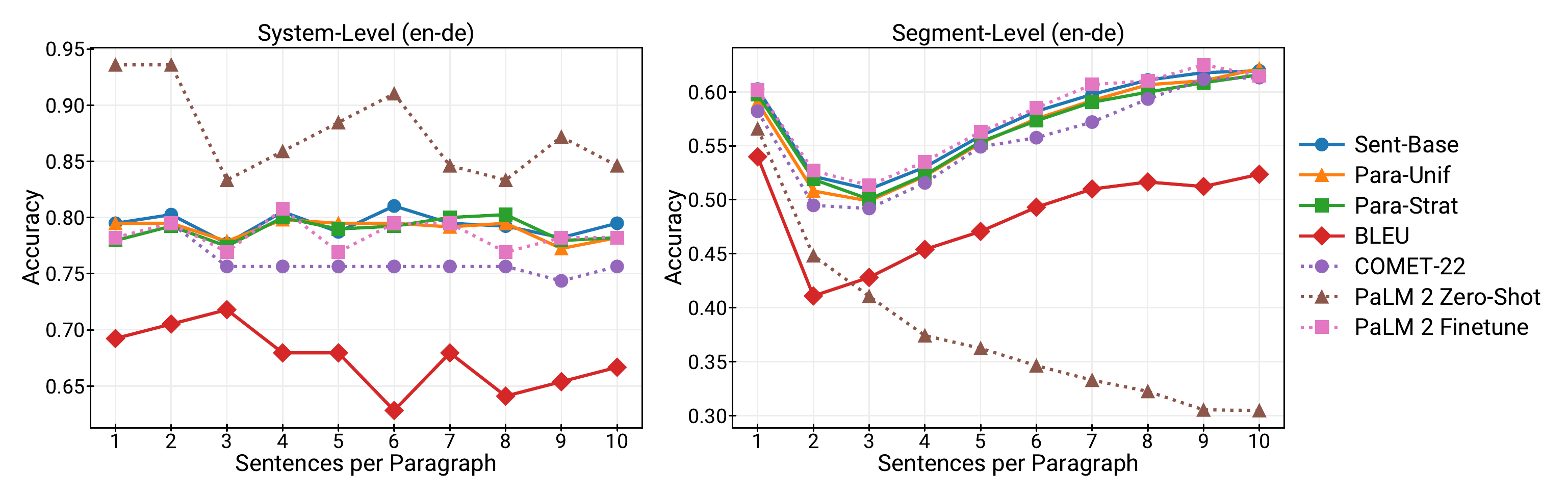}
    \includegraphics[width=\textwidth]{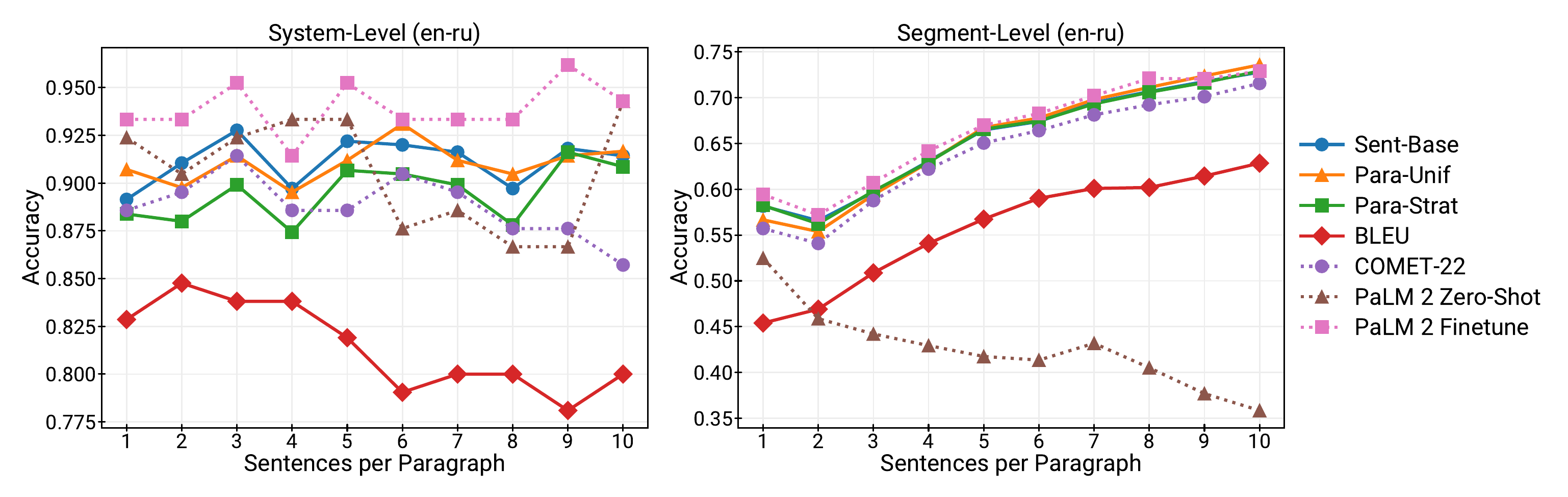}
    \caption{System- and segment-level accuracy results for the en-de and en-ru language pairs on the paragraph-level WMT'22 MQM data for different numbers of $k$ sentences per paragraph.
    In general, the system-level correlations are relatively flat and the segment-level correlations increase as the number of sentences per paragraph increases.
    BlonDE is not included because it only supports English.
    }
    \label{fig:paragraph_results_ende_enru}
\end{figure*}

%% file: figures/paragraph_results_pearson.tex
\begin{figure*}
    \centering
    \includegraphics[width=\textwidth]{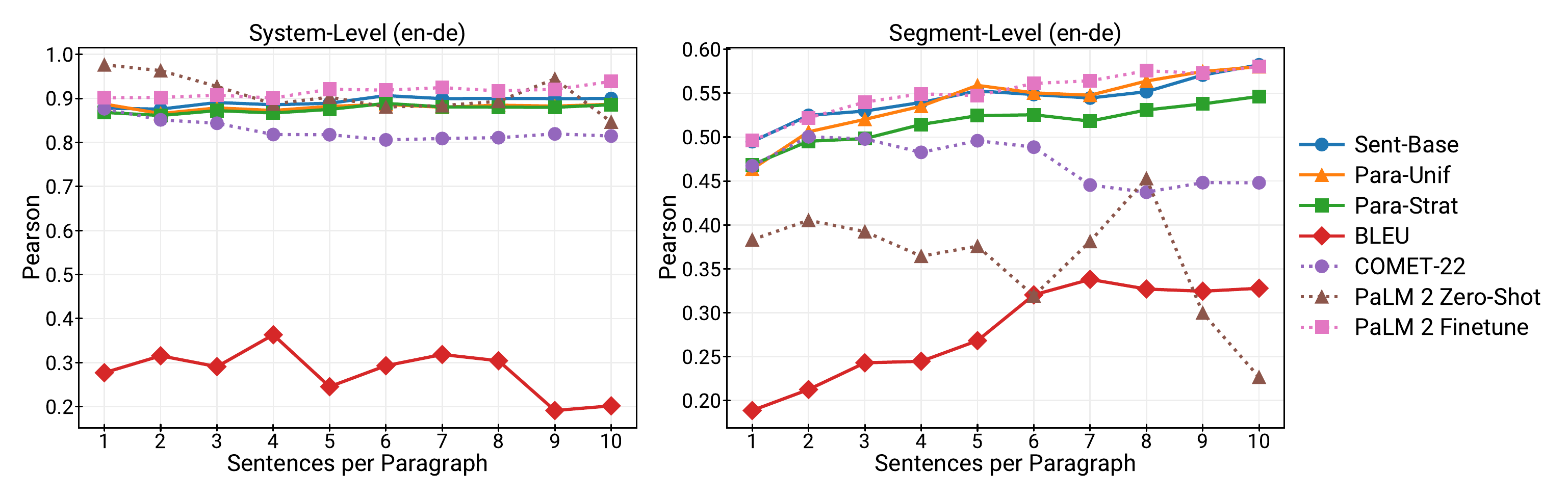}
    \includegraphics[width=\textwidth]{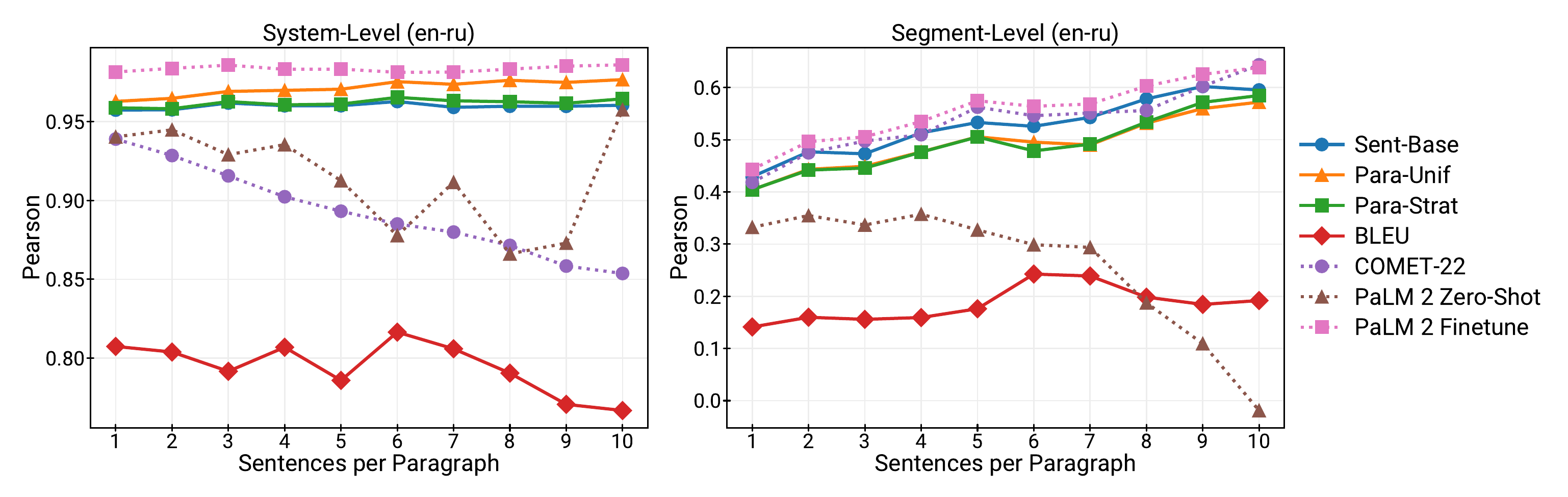}
    \includegraphics[width=\textwidth]{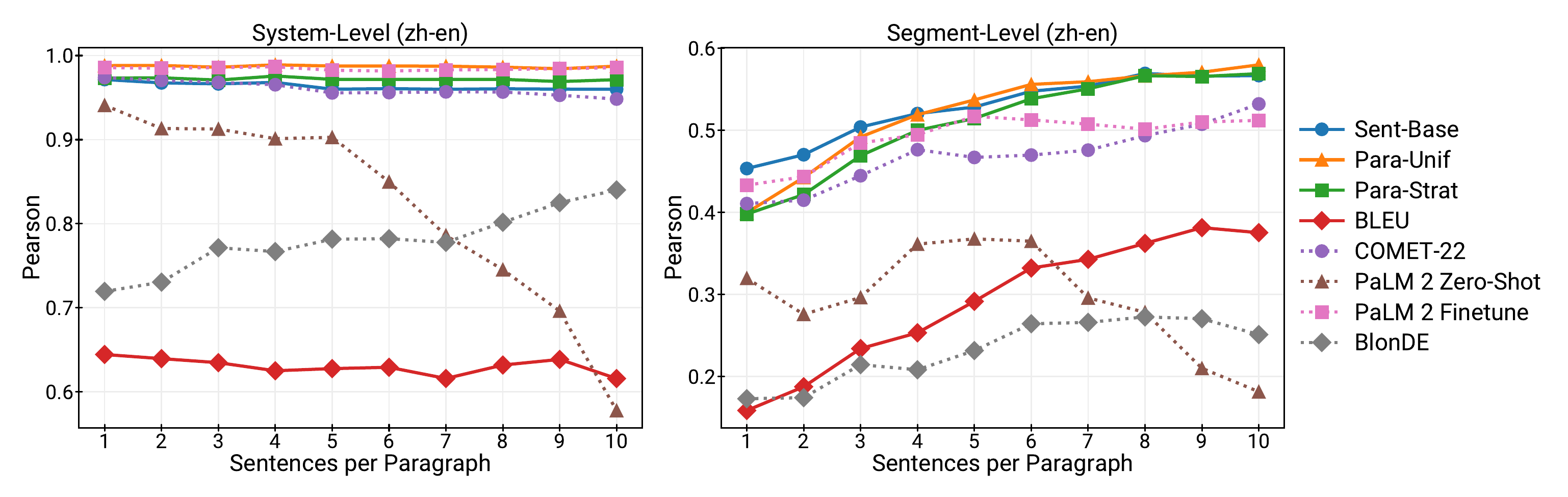}
    \caption{The system- and segment-level correlation results when using Pearson correlation follow very similar trends to those that use pairwise accuracy.
    The segment-level Pearson uses the ``no grouping'' variant from \citet{deutsch2023ties} to avoid the NaN problem that happens with the ``group-by-item'' variant, which was used in combination with pairwise accuracy in the main body of the paper.
    }
    \label{fig:paragraph_pearson}
\end{figure*}

%% file: figures/self_correlation_appendix.tex
\begin{figure}
    \centering
    \includegraphics[width=\columnwidth]{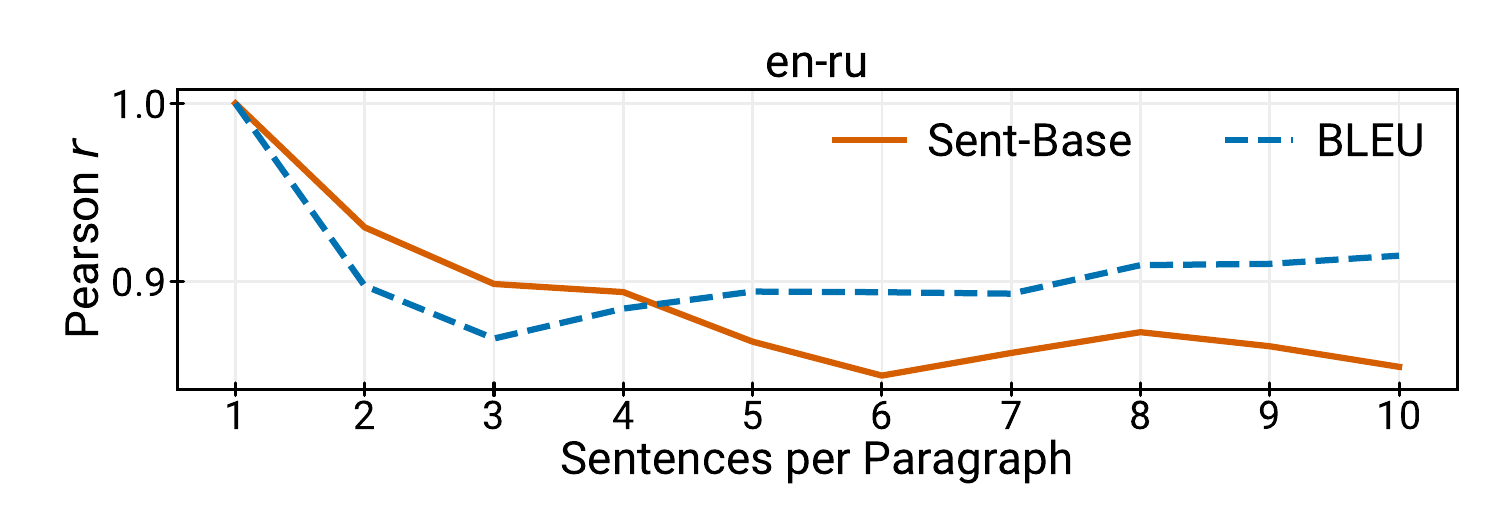}
    \includegraphics[width=\columnwidth]{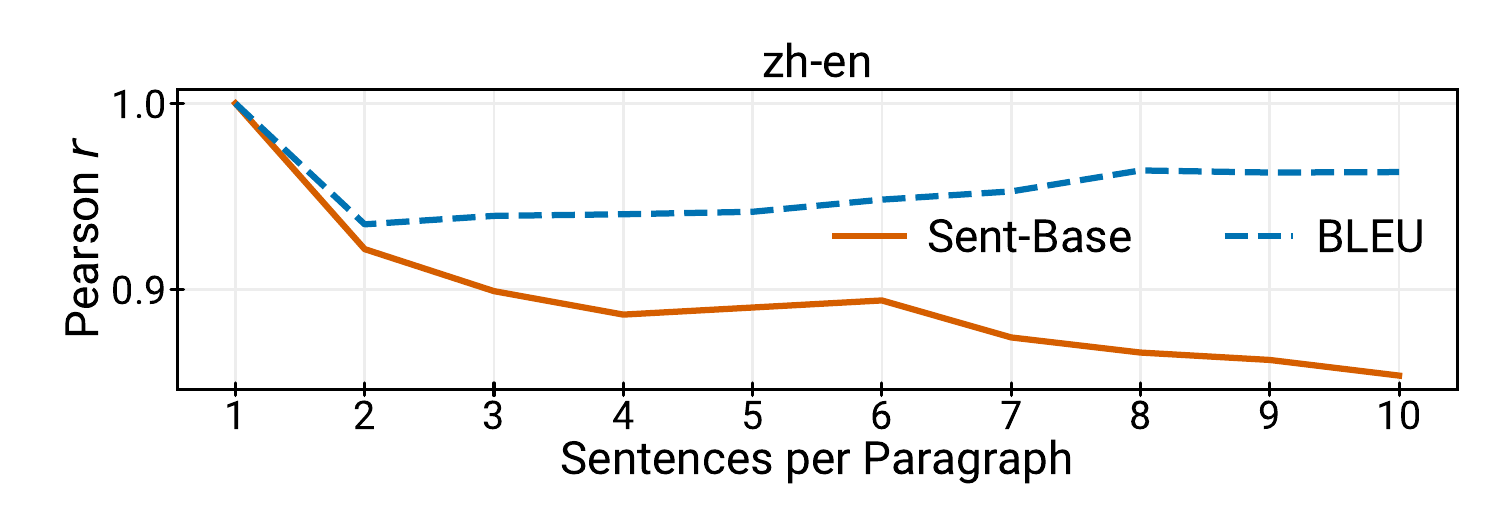}
    \caption{The correlation between metric scores for directly scoring paragraphs and averaging the score of evaluating the $k$ sentences per paragraph independently on the WMT'22 MQM data.}
    \label{fig:self_correlation_enru_zhen}
\end{figure}